\documentclass[11pt,a4paper]{article}
\usepackage[hyperref]{emnlp2020}
\usepackage{times}
\usepackage{latexsym}

\usepackage{multirow,tabularx,booktabs}
\usepackage{subfig}
\usepackage{amsmath}
\usepackage{appendix}

\usepackage{url}

\usepackage[utf8]{inputenc}
\usepackage{graphicx}

\usepackage{microtype}

\aclfinalcopy 

\usepackage[shortlabels, inline]{enumitem}

\newcommand{\mycomment}[3]{}

\usepackage{xcolor}

\newcommand{\ignore}[1]{}

\title{End-to-End Slot Alignment and Recognition for Cross-Lingual NLU}

\author{Weijia Xu\Thanks{\ Work done at Amazon AI.} \\
  University of Maryland \\
  \texttt{weijia@cs.umd.edu} \\\And
  Batool Haider \\
  Amazon AI \\
  \texttt{bhaider@amazon.com} \\\And
  Saab Mansour \\
  Amazon AI \\
  \texttt{saabm@amazon.com} \\}

\date{}

\begin{document}
\maketitle
\begin{abstract}
Natural language understanding~(NLU) in the context of goal-oriented dialog systems typically includes intent classification and slot labeling tasks. Existing methods to expand an NLU system to new languages use machine translation with slot label projection from source to the translated utterances, and thus are sensitive to projection errors. In this work, we propose a novel end-to-end model that learns to align and predict target slot labels jointly for cross-lingual transfer. We introduce MultiATIS++, a new multilingual NLU corpus that extends the Multilingual ATIS corpus to nine languages across four language families, and evaluate our method using the corpus. Results show that our method outperforms a simple label projection method using fast-align on most languages, and achieves competitive performance to the more complex, state-of-the-art projection method with only half of the training time. We release our MultiATIS++ corpus to the community to continue future research on cross-lingual NLU.
\end{abstract}

\section{Introduction}
\begin{figure*}[ht]
\centering
  \includegraphics[width=.9\linewidth,page=1, trim=4cm 5.3cm 3cm 2.5cm, clip]{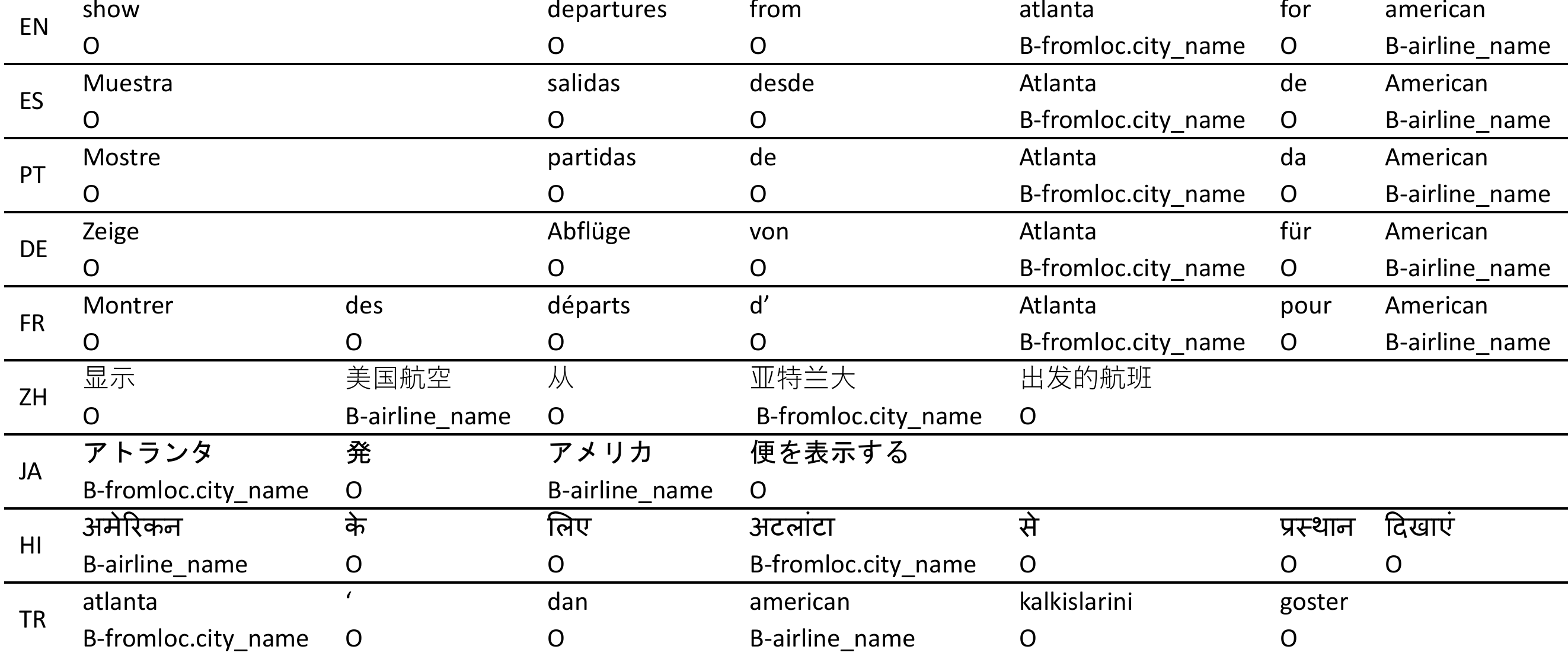}
\caption{An English training example and its translated versions in the MultiATIS++ corpus. The English utterance is manually translated to the other eight languages including Spanish~(ES), Portuguese~(PT), German~(DE), French~(FR), Chinese~(ZH), Japanese~(JA), Hindi~(HI), and Turkish~(TR). For each language, we show the utterance followed by the slot labels in the BIO format. The intent of the utterances is the \textit{flight} intent.}
\label{fig:atis_examples}
\end{figure*}

As a crucial component of goal oriented dialogue systems, natural language understanding~(NLU) is responsible for parsing an utterance into a semantic frame to identify the user's need. These semantic frames are structured by what the
user intends to do (the \textit{intent}) and the arguments of the intent ~(the \textit{slots})~\citep{Tur2010}.
Given the English example in Figure~\ref{fig:atis_examples}, we identify the intent of the utterance as ``flight'' and label the slots to extract the departure city and airline name.
Intent detection can be modeled as a sentence classification task where each utterance is labeled with an intent~$y_I$. Slot filling is typically modeled as a sequence labeling task where given the utterance~$\boldsymbol{x}_{1...n}$, each word~$x_i$ is labeled with a slot~$y_i$.

Despite the high accuracy achieved by neural models on intent detection and slot filling~\citep{Goo2018, Qin2019}, training such models on a new language requires additional efforts to collect large amounts of training data. One would consider transfer learning from high-resource to low-resource languages to minimize the efforts of data collection and annotation.
However, currently available multilingual NLU datasets~\citep{Upadhyay2018,Schuster2019} only support three languages distributed in two language families, which hinders the study of cross-lingual transfer across a broad spectrum of language distances.
In this paper, we release a new multilingual NLU corpus that contains training, development, and test data for six new languages in addition to the three languages in the Multilingual ATIS corpus~\citep{Upadhyay2018}.
The resulting corpus, namely MultiATIS++, consists in total of~37,084 training examples and~7,859 test examples covering nine languages in four language families.

\looseness=-1
Using our corpus, we explore the use of multilingual BERT encoder~\citep{Devlin2019}, machine translation~(MT), and label projection methods for multilingual training and cross-lingual transfer. Furthermore, we propose an end-to-end model for joint slot label alignment and recognition, so that it no longer relies on slot label projections using external word alignment tools \citep{Mayhew2017,Schuster2019} or engineered features \citep{Ehrmann2011,Jain2019}, which may not generalize well to low-resource languages. Our model performs soft label alignment using an attention module which is trained jointly with other model components on intent classification, slot filling, and an augmented reconstruction objective designed to improve the soft label alignment.

Experimental results show that our method uses the same amount of training time as a simple label projection approach using fast-align, while achieving significantly higher slot F1 on most languages. Furthermore, our method achieves competitive performance to the more complex, state-of-the-art label projection method that uses linguistic features to improve projection quality, while using half of the training time. Finally, our results show different trends when comparing various cross-lingual transfer methods on different languages, which emphasizes the need to evaluate cross-lingual transfer methods on a diverse set of languages to fully illustrate the strengths and weaknesses of each method.

\section{Related Work}
Cross-lingual transfer learning has been studied on a variety of sequence tagging tasks including part-of-speech tagging~\citep{Yarowsky2001,Tackstrom2013,Plank2018}, named entity recognition~\citep{Zirikly2015,Tsai2016,Xie2018}, and natural language understanding~\citep{HeDHT2013,Upadhyay2018,Schuster2019}.
Existing methods can be roughly categorized into two categories: transfer through multilingual models and transfer through machine translation.

\paragraph{Transfer via Multilingual Models}
For closely related languages with similar alphabets, it is beneficial to train a multilingual model with shared character encoder to learn common character-based features~\citep{Yang2017,Lin2018}. However, such techniques are less effective when applied to dissimilar languages that lack common lexical features. \citet{Chen2018, Chen2019} focus on the multi-source transfer scenario and apply adversarial training to extract language-invariant features shared by source languages.
Recent advances on cross-lingual representations have enabled transfer between dissimilar languages. Representations from multilingual neural machine translation~(NMT) encoders have been shown to be effective for cross-lingual text classification~\citep{Eriguchi2018,Yu2018,Singla2018}. 
In this work, we use multilingual BERT~\citep{Devlin2019}, an unsupervised cross-lingual language model trained on monolingual texts from a wide range of languages and has been shown to provide powerful sentence representations that lead to promising performance for zero-resource cross-lingual language understanding tasks~\citep{Pires2019}. We leave the use of other recently proposed cross-lingual language models such as XLM-R~\citep{Conneau20XLMR} to future work.

\paragraph{Transfer via Machine Translation}
requires translating the source language training data into the target language or translating the target language test data into the source language. Despite its empirical success on cross-lingual text classification tasks~\citep{Wan2009}, it faces a challenging problem on the sequence tagging tasks: labels on the source language sentences need to be projected to the translated sentences. Most of the prior work relies on unsupervised word alignment from statistical MT~\citep{Yarowsky2001,Shah2010,Ni2017} or attention weights from NMT models~\citep{Schuster2019}. Other heuristic approaches include matching tokens based on their surface forms~\citep{Feng2004,Samy2005,Ehrmann2011} and more complex projection approaches that combine linguistic features with information from the MT systems~\citep{Jain2019}. 
By contrast, our method does not rely on external word alignment or linguistic features, but models label projection through an attention module that can be jointly trained with other model components on the machine translated data.

\begin{table*}[ht]
\centering
\begin{tabular}{lrrrrrrrr}
\toprule
\multirow{2}{*}{Language} & \multicolumn{3}{c}{Utterances} & \multicolumn{3}{c}{Tokens} & \multirow{2}{*}{Intents} & \multirow{2}{*}{Slots} \\
 & train & dev & test & train & dev	& test & & \\
\hline
English	& 4488	& 490	& 893	& 50755	& 5445	& 9164	& 18	& 84 \\
Spanish	& 4488	& 490	& 893	& 55197	& 5927	& 10338	& 18	& 84 \\
Portuguese	& 4488	& 490	& 893	& 55052	& 5909	& 10228	& 18	& 84 \\
German	& 4488	& 490	& 893	& 51111	& 5517	& 9383	& 18	& 84 \\
French	& 4488	& 490	& 893	& 55909	& 5769	& 10511	& 18	& 84 \\
Chinese	& 4488	& 490	& 893	& 88194	& 9652	& 16710	& 18	& 84 \\
Japanese	& 4488	& 490	& 893	& 133890& 14416	& 25939	& 18	& 84 \\
Hindi	& 1440	& 160	& 893	& 16422	& 1753	& 9755	& 17	& 75 \\
Turkish	& 578	& 60	& 715	& 6132	& 686	& 7683	& 17	& 71 \\
\hline
\end{tabular}
\caption{Data statistics for the MultiATIS++ corpus. The number of \textit{utterances} and \textit{tokens}~(characters for Chinese and Japanese) are provided for the training~(train), development~(dev), and test sets for each of the nine languages. The total number of \textit{intents} and \textit{slots} (before adding the BIO tags) are also given.}
\label{tab:atis}
\end{table*}

\section{Data}
One of the most popular datasets for multilingual NLU is the ATIS dataset~\cite{Price1990} and its multilingual extension~\cite{Upadhyay2018}.
The ATIS dataset is created by asking each participant to interact with an agent (who has access to a database) to solve a given air travel planning problem. 
\citet{Upadhyay2018} extend the English ATIS to Hindi and Turkish by manually translating and annotating a subset of the training and test data via crowdsourcing.\footnote{\url{https://catalog.ldc.upenn.edu/LDC2019T04}}

\looseness=-1
To facilitate study on cross-lingual transfer across a broader spectrum of language distances, we create the MultiATIS++ corpus by extending both the training and test set of the English ATIS corpus to six additional languages.\footnote{\url{https://github.com/amazon-research/multiatis}} The resulting corpus covers nine languages in four different language families including Indo-European~(English, Spanish, German, French, Portuguese, and Hindi), Sino-Tibetan~(Chinese), Japonic~(Japanese), and Altaic~(Turkish).

For each new language, we hire professional native translators
to translate the English utterances and annotate the slots at the same time.
When translating, we ask the translators to preserve the spoken modality phenomena (e.g. hesitations and word repetitions) and style (e.g. degree of formality) of the original English sentences, so that it is closer to the real scenaria.
To get the slot labels for the translated utterances, we ask the translators to tag the segments in the translated utterances that are aligned to the corresponding English segments. 
Finally, we tokenize the translated utterances and BIO tag the tokens. For quality control, we ask a third-party to perform several rounds of qualification checks until no issues are reported.

We show an English training example and its translated versions in the other eight languages in Figure~\ref{fig:atis_examples}
and report the data statistics in Table~\ref{tab:atis}. We split the English training and development sets randomly and keep the same split for all the other languages except for Hindi and Turkish from the Multilingual ATIS corpus.
Note that the Hindi and Turkish portions of the data are smaller than the other languages, covering only a subset of the intent and slot types.

\section{Cross-Lingual NLU}
\begin{figure*}[ht]
\centering
  \includegraphics[width=.7\linewidth,trim=0cm 0.4cm 0cm 0.3cm, clip]{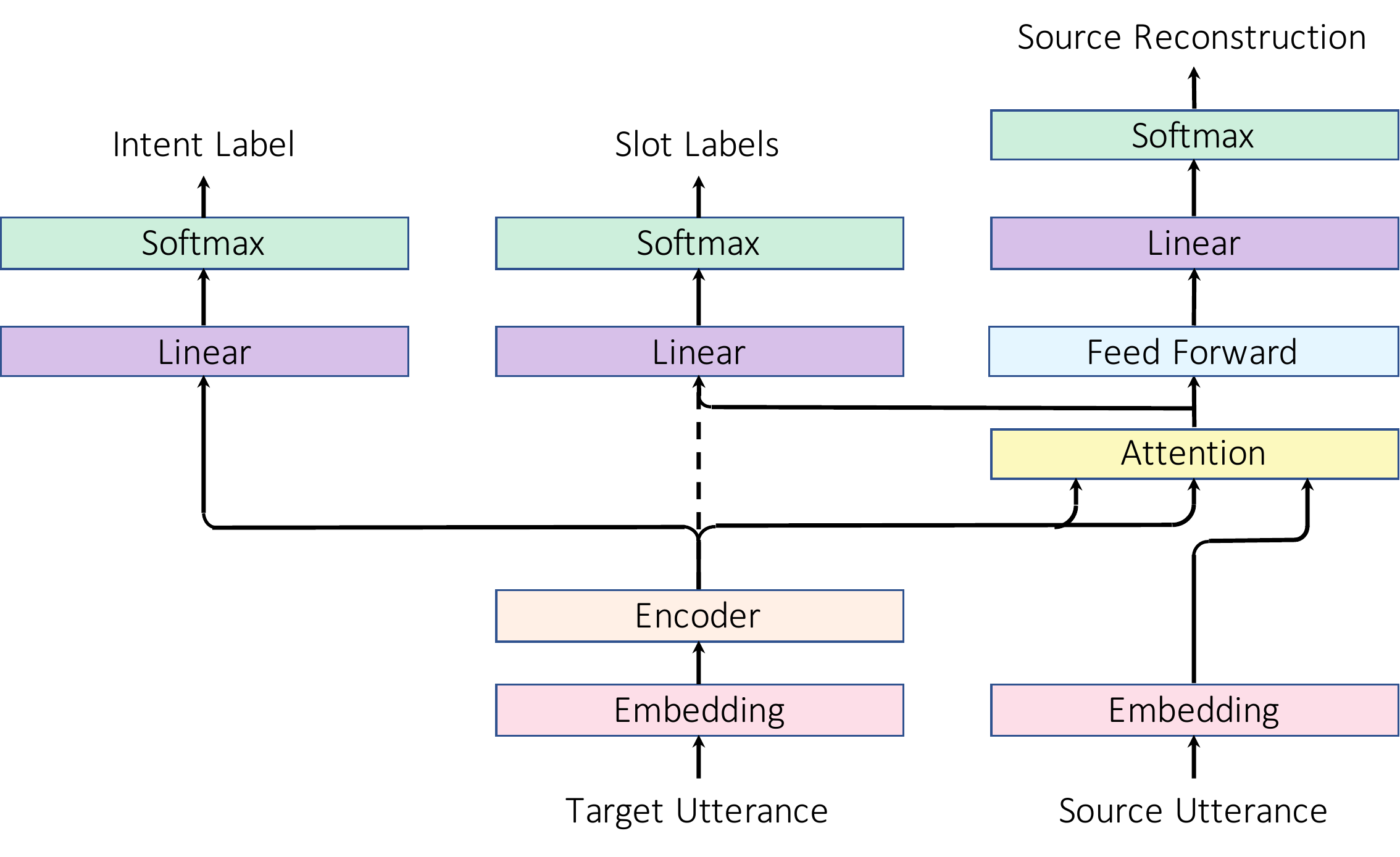}
\caption{Architecture of our soft-alignment model for end-to-end slot alignment and recognition. The model is trained without external label projection: it learns to soft-align the representations of the target utterance to the source slot labels and predict the intent and slot labels jointly. The dotted line denotes the path during inference, where we directly connect the encoder module to the intent and slot classification layer to make predictions on the target utterance.}
  \label{fig:architecture}
\end{figure*}

\subsection{Joint Intent Detection and Slot Filling}
Following~\citet{Liu2016}, we model intent detection and slot filling jointly.
We add a special classification token~$x_0$ at the beginning of the input sequence~$\boldsymbol{x} = (x_1, x_2, ..., x_T)$ of length~$T$ following~\citet{Devlin2019}. Next, an encoder~$\Theta_{enc}$ is used to produce a sequence of contextualized representations~$\boldsymbol{h}_{0...T}$ given the input sequence
\[
    \boldsymbol{h}_{0...T} = \Theta_{enc} (x_0, x_1, ..., x_T)
\]
For intent detection, we take the representation~$\boldsymbol{h}_0$ corresponding to $x_0$
as the sequence representation and apply a linear transformation and a softmax function to predict the intent probability
\[
    p_\text{intent}(\cdot | \boldsymbol{x}) = \text{softmax}(\boldsymbol{W}^I \boldsymbol{h}_0 + \boldsymbol{b}^I)
\]
For slot filling, we compute the probability for each slot using the representations~$\boldsymbol{h}_{1...T}$
\[
    p_{\text{slot}_i} (\cdot | \boldsymbol{x}) = \text{softmax}(\boldsymbol{W}^S \boldsymbol{h}_i + \boldsymbol{b}^S)
\]

We explore two different encoder models: 
\begin{itemize}
    \item {\bf LSTM:} We use the concatenation of the forward and backward hidden states of a bidirectional LSTM~\citep{Schuster1997} as the encoder representations. We initialize the encoder and embeddings randomly.
    \item {\bf Multilingual BERT:} We use the multilingual BERT~\citep{Devlin2019} pre-trained in an unsupervised way on the concatenation of monolingual corpora from~104 languages. We take the hidden states from the top layer as the encoder representations and fine-tune the model on the NLU data.
\end{itemize}

\subsection{Problems in Slot Label Projection}
Past work has shown the effectiveness of using MT systems to boost the performance of cross-lingual NLU~\cite{Schuster2019}. More specifically, one first translates the English training data to the target language using an MT system, and then projects the slot labels from English to the target language. Prior work projects the slot labels using word alignments from statistical MT models~\citep{Yarowsky2001} or attention weights from neural MT models~\citep{Schuster2019}. 
The final performance on the target language highly depends on the quality of the slot projection.
\citet{Jain2019} show that improving the quality of projection leads to significant improvements in the target performance on cross-lingual named entity recognition. However, the improvements come at the cost of much more complex and expensive projection process using engineered features. 

\looseness=-1
To address the above issues, we propose a soft-alignment model that performs end-to-end slot alignment and recognition using an additional attention module~(Figure~\ref{fig:architecture}), so that it requires no external slot projection process. Furthermore, we show that the soft slot alignment can be strengthened by building it on top of strong encoder representations from multilingual BERT.

\subsection{End-to-End Slot Alignment and Recognition via Attention}
\label{sec:learning_to_align}

Given a source utterance~$\boldsymbol{s}_{1...S}$ of length~$S$ and its translation~$\boldsymbol{t}_{1...T}$ of length~$T$ in the target language, the model learns to predict the target slot labels and soft-align it with the source labels via attention. First, it encodes the source utterance into a sequence of embeddings~$\boldsymbol{e}^{(src)}_{1...S}$ and then encodes the translation~$\boldsymbol{t}_{0...T}$~($t_0$ is inserted as the classification token) into a sequence of contextualized representations~$\boldsymbol{h}^{(tgt)}_{0...T} = \Theta_{enc}(\boldsymbol{t}_{0...T})$, where~$\Theta_{enc}$ represents the encoder. For intent classification, we assume that the translated utterance has the same intent as the source utterance. Thus we compute the intent probabilities using the representation~$\boldsymbol{h}_0^{(tgt)}$
\[
    p_\text{intent}(\cdot | \boldsymbol{t}) = \text{softmax}(\boldsymbol{W}^I \boldsymbol{h}_0^{(tgt)} + \boldsymbol{b}^I)
\]
and the intent classification loss given the intent label~$y_I^{(src)}$ on the source utterance
\[
    \mathcal{L}_\text{intent} = -\log p_\text{intent}(y_I^{(src)} | \boldsymbol{t}) 
\]

For slot filling, we introduce an attention module to connect the source slot labels~$\boldsymbol{y}^{(src)}_{1...S}$ with the target sequence~$\boldsymbol{t}_{1...T}$.
First, we compute the hidden state at each source position as a weighted average of the target representations
\[
    \boldsymbol{z}_i = \sum_{j=1}^T a_{ij} \boldsymbol{h}_j^{(tgt)}
\]
where~$\boldsymbol{z}_i$ is the hidden state at source position~$i$, and~$a_{ij}$ is the attention weights between the source word~$s_i$ and translation word~$t_j$. To compute the weights~$a_{ij}$, we first linearly project the query vector~$\boldsymbol{e}_i^{(src)}$ and the key vector~$\boldsymbol{h}_j^{(tgt)}$ with learnable parameters to~$d$ dimensions. We then perform the scaled dot-product attention on the projected query and key vectors
\[
    \boldsymbol{a}_i = \text{softmax}\left(\frac{ (\boldsymbol{e}_i^{(src)} \boldsymbol{W}^Q) (\boldsymbol{h}^{(tgt)} \boldsymbol{W}^K)^\mathsf{T} }{\sqrt{d} \tau}\right)
\]
where the projections~$\boldsymbol{W}^Q$ and~$\boldsymbol{W}^K$ are parameter matrices, and~$\tau$ is a hyperparameter that controls the temperature of the softmax function.

Next, we compute the slot probabilities at the source position~$i$ using the hidden state~$\boldsymbol{z}_i$
\[
    p_{\text{slot}_i} (\cdot | s_i, \boldsymbol{t}) = \text{softmax}(\boldsymbol{W}^S \boldsymbol{z}_i + \boldsymbol{b}^S)
\]
and the slot filling loss given the slot labels~$y^{(src)}_{1...S}$ on the source utterance
\[
    \mathcal{L}_\text{slot} = -\sum_{i=1}^S \log p_{\text{slot}_i}(y_i^{(src)} | s_i, \boldsymbol{t}) 
\]

\looseness=-1
In addition, to improve the attention module to better align the source and target utterances, we add a reconstruction module consisting of a position-wise feed-forward and a linear output layer\footnote{We tie the output weights~$\boldsymbol{W}^R$ with BERT embeddings.} to recover the source utterance using the attention outputs. We compute the probability distribution over the source vocabulary at position~$i$ as
\begin{equation}
\begin{split}
    & p_{\text{rec}_i} (\cdot | s_i, \boldsymbol{t}) = \text{softmax}(\boldsymbol{W}^R \boldsymbol{\tilde{z}}_i + \boldsymbol{b}^R) \\
    & \boldsymbol{\tilde{z}}_i = \text{FeedForward}(\boldsymbol{z}_i)
\end{split}
\end{equation}
and the reconstruction loss as
\begin{equation}
    \mathcal{L}_\text{rec} = -\sum_{i=1}^S \log p_{\text{rec}_i}(s_i | s_i, \boldsymbol{t})
\end{equation}
The final training loss is~$\mathcal{L} = \mathcal{L}_\text{intent} + \mathcal{L}_\text{slot} + \mathcal{L}_\text{rec}$. 

\looseness=-1
Empirically, we find it beneficial to train the model jointly on the machine translated target data using the objective~$\mathcal{L}$ and the source data using the supervised objective.

The attention and reconstruction modules are only used during training. During inference, we directly feed the encoder representations~$\boldsymbol{h}^{(tgt)}_{0...T}$ of the target language utterance to the intent and slot classification layers
\begin{equation}
\begin{split}
    & p_\text{intent}(\cdot | \boldsymbol{t}) = \text{softmax}(\boldsymbol{W}^I \boldsymbol{h}_0^{(tgt)} + \boldsymbol{b}^I) \\
    & p_{\text{slot}_i} (\cdot | \boldsymbol{t}) = \text{softmax}(\boldsymbol{W}^S \boldsymbol{h}_i^{(tgt)} + \boldsymbol{b}^S) \\
\end{split}
\end{equation}
\begin{table*}[ht]
\centering
\begin{tabular}{llrrrrrrrrr}
\toprule
\multicolumn{2}{l}{\bf Intent acc.} & {en} & {es} & {de} & {zh} & {ja} & {pt} & {fr} & {hi} & {tr} \\
\midrule
\multirow{2}{*}{Target only} & {LSTM} & {96.08} & {93.04} & {94.02} & {92.50} & {91.18} & {92.70} & {94.71} & {84.46} & {81.12} \\
{} & {BERT} & {\textbf{97.20}} & {\bf 96.44} & {\bf 96.73} & {\bf 95.52} & {95.54} & \textbf{{96.71}} & \textbf{{97.38}} & {90.50} & {87.10} \\
\multirow{2}{*}{Multilingual} & {LSTM} & {95.45} & {94.09} & {95.05} & {93.42} & {92.90} & {94.02} & {94.80} & {87.79} & {85.43} \\
{} & {BERT} & \textbf{{97.20}} & \textbf{{96.77}} & \textbf{{96.86}} & \textbf{{95.54}} & \textbf{{96.44}} & {96.48} & {\bf 97.24} & \textbf{{92.70}} & \textbf{{92.20}} \\
\bottomrule
\multicolumn{2}{l}{\bf Slot F1} & {en} & {es} & {de} & {zh} & {ja} & {pt} & {fr} & {hi} & {tr} \\
\midrule
\multirow{2}{*}{Target only} & {LSTM} & {94.71} & {75.89} & {91.44} & {90.84} & {88.80} & {88.43} & {85.93} & {74.93} & {64.43} \\
{} & {BERT} & {95.57} & {86.58} & {\bf 94.98} & {\bf 93.52} & {91.40} & {91.35} & {89.14} & {82.36} & {75.21} \\
\multirow{2}{*}{Multilingual} & {LSTM} & {94.75} & {84.11} & {92.00} & {90.76} & {88.55} & {88.79} & {87.96} & {77.34} & {77.25} \\
{} & {BERT} & \textbf{{95.90}} & \textbf{{87.95}} & \textbf{{95.00}} & \textbf{{93.67}} & \textbf{{92.04}} & \textbf{{91.96}} & \textbf{{90.39}} & \textbf{{86.73}} & \textbf{{86.04}} \\
\bottomrule
\end{tabular}
\caption{Results on MultiATIS++ using full training data and the standard supervised objective averaged over 5 runs. The \textit{Target only} models are trained only on the target language training data. The \textit{Multilingual} models are trained on the concatenation of training data from all languages.}
\label{tab:multi_results}
\end{table*}
\begin{table*}
\centering
\begin{tabular}{llrrrrrrrr}
\toprule
\multicolumn{2}{l}{\bf Intent acc.} & {es} & {de} & {zh} & {ja} & {pt} & {fr} & {hi} & {tr} \\
\midrule
\multirow{2}{*}{No MT} & {LSTM} & {64.82} & {64.77} & {59.69} & {65.40} & {65.15} & {69.92} & {60.11} & {63.64} \\
{} & {BERT} & {96.35} & {95.27} & {86.27} & {79.42} & {94.96} & {95.92} & {80.96} & {69.59} \\
\multirow{2}{*}{MT+fast-align} & {LSTM} & {95.36} & {94.02} & {92.70} & {77.96} & {94.20} & {94.76} & {88.71} & {87.72} \\
{} & {BERT} & {\bf 97.02} & {\textbf{96.77}} & \textbf{{96.10}} & \textbf{{88.82}} & {96.55} & {96.89} & \textbf{{93.12}} & \textbf{{93.77}} \\
\multirow{2}{*}{MT+TMP} & {LSTM} & {95.32} & {93.03} & {90.41} & {84.43} & {94.36} & {94.15} & {88.67} & {87.47} \\
{} & {BERT} & {\bf 97.00} & {96.01} & {95.16} & {\bf 88.51} & {96.46} &	{97.04} & {92.41} & {\bf 93.74} \\
{\bf MT+soft-align} & {BERT} & {\textbf{97.20}} & {\bf 96.66} & {\bf 95.99} & {\bf 88.33} & \textbf{{96.78}} & \textbf{{97.49}} & {92.81} & {\bf 93.71} \\
\bottomrule
\multicolumn{2}{l}{\bf Slot F1} & {es} & {de} & {zh} & {ja} & {pt} & {fr} & {hi} & {tr} \\
\midrule
\multirow{2}{*}{No MT} & {LSTM} & {27.98} & {32.96} & {1.60} & {2.71} & {25.52} & {29.70} & {2.26} & {26.56} \\
{} & {BERT} & {74.98} & {82.61} & {62.27} & {35.75} & {74.05} & {75.71} & {31.21} & {23.75} \\
\multirow{2}{*}{MT+fast-align} & {LSTM} & {76.30} & {83.83} & {78.61} & {70.23} & {76.28} & {64.37} & {60.02} & {21.53} \\
{} & {BERT} & {79.18} & {87.21} & {81.82} & {79.53} & {78.26} & {70.18} & {69.42} & {23.61} \\
\multirow{2}{*}{MT+TMP} & {LSTM} & {79.19} & {84.99} & {82.84} & {71.98} & {79.44} & {77.52} & {67.70} & {40.14} \\
{} & {BERT} & {\bf 83.98} & {87.54} & {\bf 85.05} & {\bf 82.60} & {\bf 81.73} & {\bf 79.80} & {77.24} & {44.80} \\
{\bf MT+soft-align} & {BERT} & {76.42} & \textbf{{89.00}} & {83.25} & {79.10} & {76.30} & {\bf 79.64} & {\bf 78.56} & \textbf{{61.70}} \\
\bottomrule
\end{tabular}
\caption{Zero-shot results on MultiATIS++ averaged over 5 runs. The \textit{No MT} rows are models trained only on the English data. The \textit{MT+fast-align} rows correspond to models trained on the English and machine translated data with automatically projected slot labels using \textit{fast-align}, and \textit{MT+TMP} correspond to Translate-Match-Project~\citep{Jain2019}. The \textit{MT+soft-align} row is the model trained on the English and machine translated data using our soft-alignment method.}
\label{tab:zero_results}
\end{table*}

\section{Multilingual NLU}
\label{sec:cross_lingual_exp}
In our first set of experiments, we explore using pre-trained multilingual BERT encoder for multilingual NLU. We compare the following training strategies to leverage the full supervised training data:
\begin{itemize}
    \item {\bf Target only:} Train each model on the target language training data using the standard supervised objective.
    \item {\bf Multilingual:} Train a model on the concatenation of training data from all languages using the standard supervised objective.
\end{itemize}

\paragraph{Setup}
We train the models using the Adam optimizer~\cite{Kingma2015} for 20 epochs and select the model that performs the best on the development set (details in Appendix~\ref{appendix:model}). Following \cite{Goo2018}, we use intent accuracy and slot F1 as evaluation metrics.

\paragraph{Results}
Table~\ref{tab:multi_results} shows the results using full training data and the supervised objective. First, we compare LSTM and BERT models trained on the target language data only. Multilingual BERT encoder brings significant\footnote{All mentions of statistical significance are based on paired Student's t-test with $p <0.05$.} improvements of~1--6\% on intent accuracy and~1--11\% on slot F1. The largest improvements are on the two low-resource languages: Hindi and Turkish. Multilingual training on all languages brings further improvements on Hindi and Turkish: it improves intent accuracy by~2--5\% and slot F1 by~4--11\% for both LSTM and BERT models.

\paragraph{Comparison with SOTA}
On English ATIS, \citet{Qin2019} report 97.5\% intent accuracy and 96.1\% slot F1 when using BERT with their proposed stack-propagation architecture. This is comparable to our \textit{target only with BERT} scores in Table~\ref{tab:multi_results}. On multilingual ATIS, \citet{Upadhyay2018} report slot F1 of 80.6\% on Hindi and 78.9\% on Turkish using bilingual training. Our multilingual BERT model achieves higher F1 by +6.1\% on Hindi and +7.1\% on Turkish.

\section{Cross-Lingual Transfer}
In this section, we compare the following methods for cross-lingual transfer where we only use the English training data and a small amount~(few-shot) or no~(zero-shot) training data from the target language: 
\begin{itemize}
    \item {\bf No MT:} Train the models only on the English training data without machine translating them to the target language.
    \item {\bf MT+fast-align:} Use MT systems to translate the English data to the target language and project the slot labels using word alignment from \textit{fast-align}.\footnote{\url{https://github.com/clab/fast_align}}
    \item {\bf MT+TMP:} Use MT to translate the English data to the target language and the Translate-Match-Project method~\citep{Jain2019} to project the slot labels.
    \item {\bf MT+soft-align:} Use MT to translate the English data to the target language and our soft-alignment method described in Section~\ref{sec:learning_to_align}.
\end{itemize}

For all three MT-based approaches, we use AWS Translate for automatic translation and perform multilingual training on the English and machine translated data. We adopt the same setup as the previous section, except that we select the model at the last epoch as we assume no access to the development set from the target language in this setting. For the attention module in our soft-alignment model, we set the temperature~$\tau = 0.1$.

\begin{table}[ht]
\centering
\begin{tabular}{llr}
\toprule
\multicolumn{2}{l}{} & {Time (mins)} \\
\midrule
\multirow{2}{*}{No MT} & {LSTM} & 18  \\
{} & {BERT} & 65 \\
\multirow{2}{*}{MT+fast-align} & {LSTM} & 232 \\
{} & {BERT} & 317  \\
\multirow{2}{*}{MT+TMP} & {LSTM} & 626 \\
{} & {BERT} & 719  \\
{MT+soft-align} & {BERT} & 352 \\
\bottomrule
\end{tabular}
\caption{Total training time of each method on all languages for the zero-shot transfer experiments (including the time for machine translation, label projection, and model training). \textit{MT+TMP} requires a time-consuming label projection process and thus takes double the time as our soft-alignment method.}
\label{tab:zeroshot_runtime}
\end{table}

\subsection{Zero-Shot Results}
\looseness=-1
Table~\ref{tab:zero_results} shows the results on zero-shot cross-lingual transfer.
First, we study the impact of using multilingual BERT for MT-based transfer approaches. For \textit{MT+fast-align}, BERT boosts the performance over LSTM by large margins:~14--32\% on intent accuracy and~29--61\% on slot F1 on all languages except for Turkish \---\ a dissimilar language to English. For both LSTM and BERT models, using \textit{MT+fast-align} brings significant improvements over their counterparts without MT on intent accuracy \---\ improvements of~13--33\% when using LSTM and~1--24\% when using BERT. However, we observe different trends on slot F1 for different languages. For example, when using BERT, \textit{MT+fast-align} improves slot F1 by~20--44\% on Chinese, Japanese, and Hindi over BERT without MT, but hurts by around~6\% on French.\footnote{The difference is significant with~$p < 0.05$.} This is possibly because that the mBERT representations of French are of high quality, which leads to relatively high slot F1 without MT, thus adding more noisy data via \textit{MT+fast-align} does more harm than good. This indicates that, while training directly on the target language data is beneficial especially for languages dissimilar to English, the noisy projection of the slot labels could also bring harm to the model.

\begin{figure*}
	\centering
    \subfloat[Intent accuracy on French]{{\includegraphics[width=0.46\textwidth]{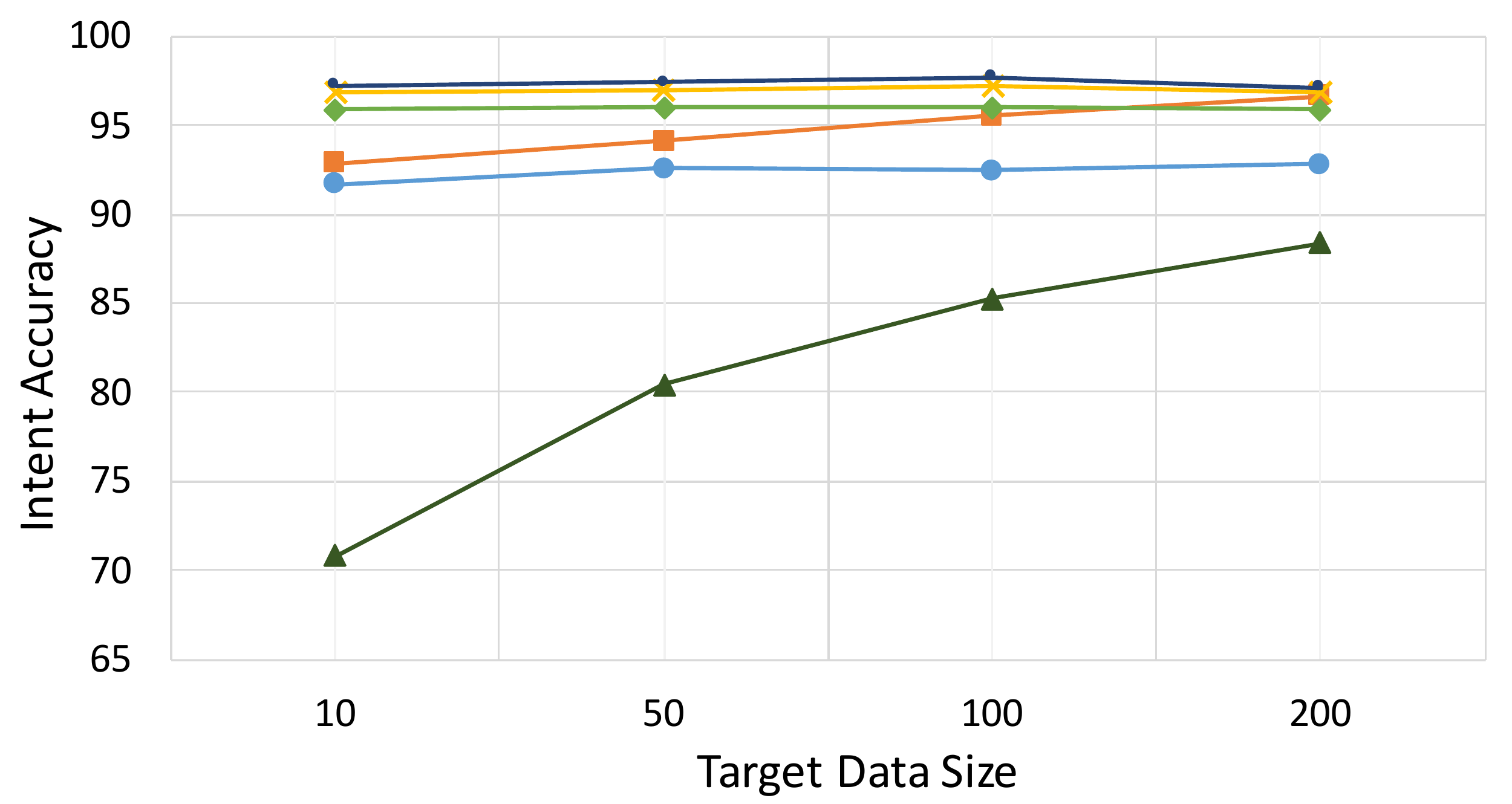} }}\hspace{1em}
    \subfloat[Intent accuracy on Chinese]{{\includegraphics[width=0.46\textwidth]{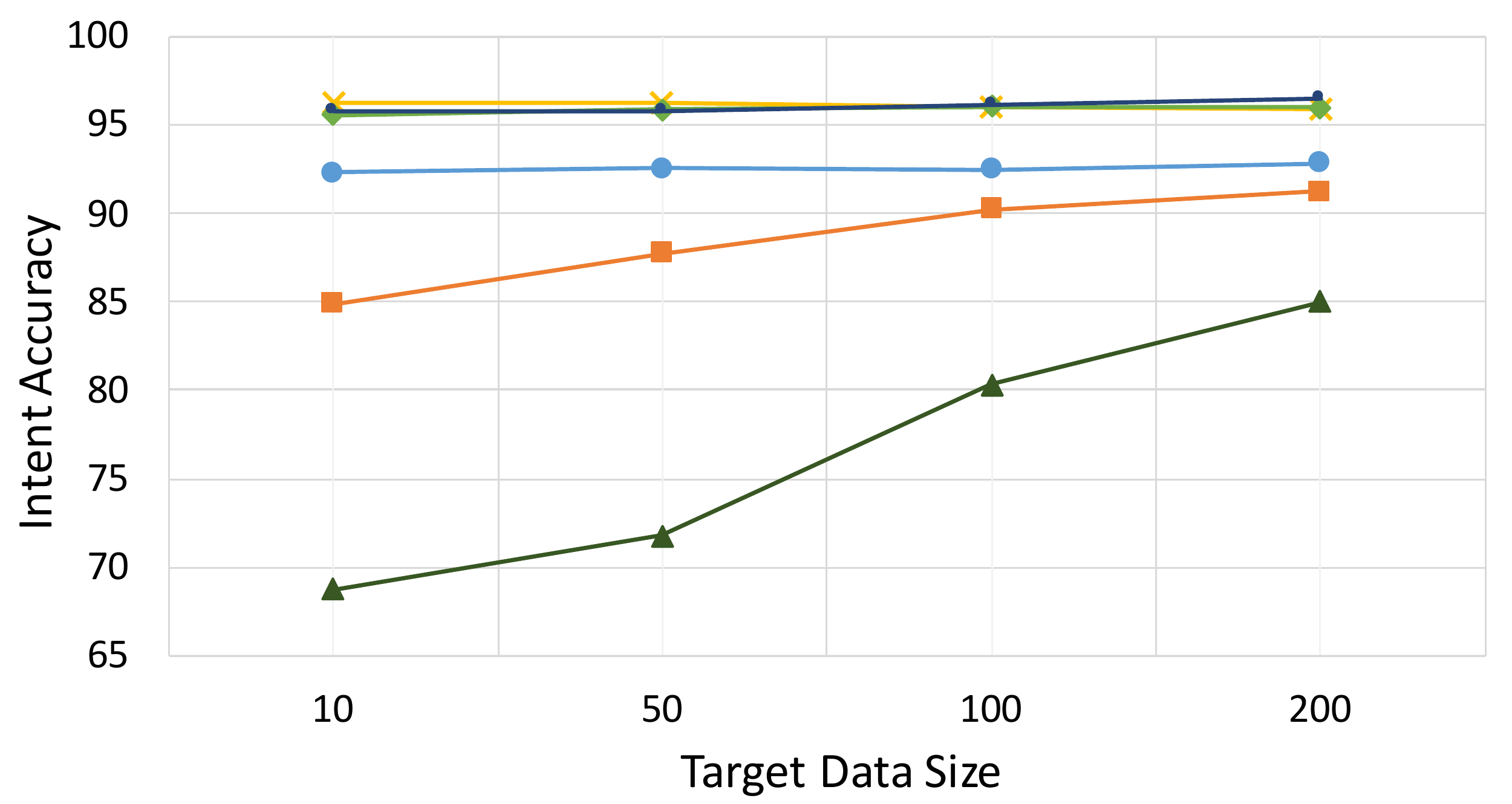} }} \\
    \subfloat[Slot F1 on French]{{\includegraphics[width=0.46\textwidth]{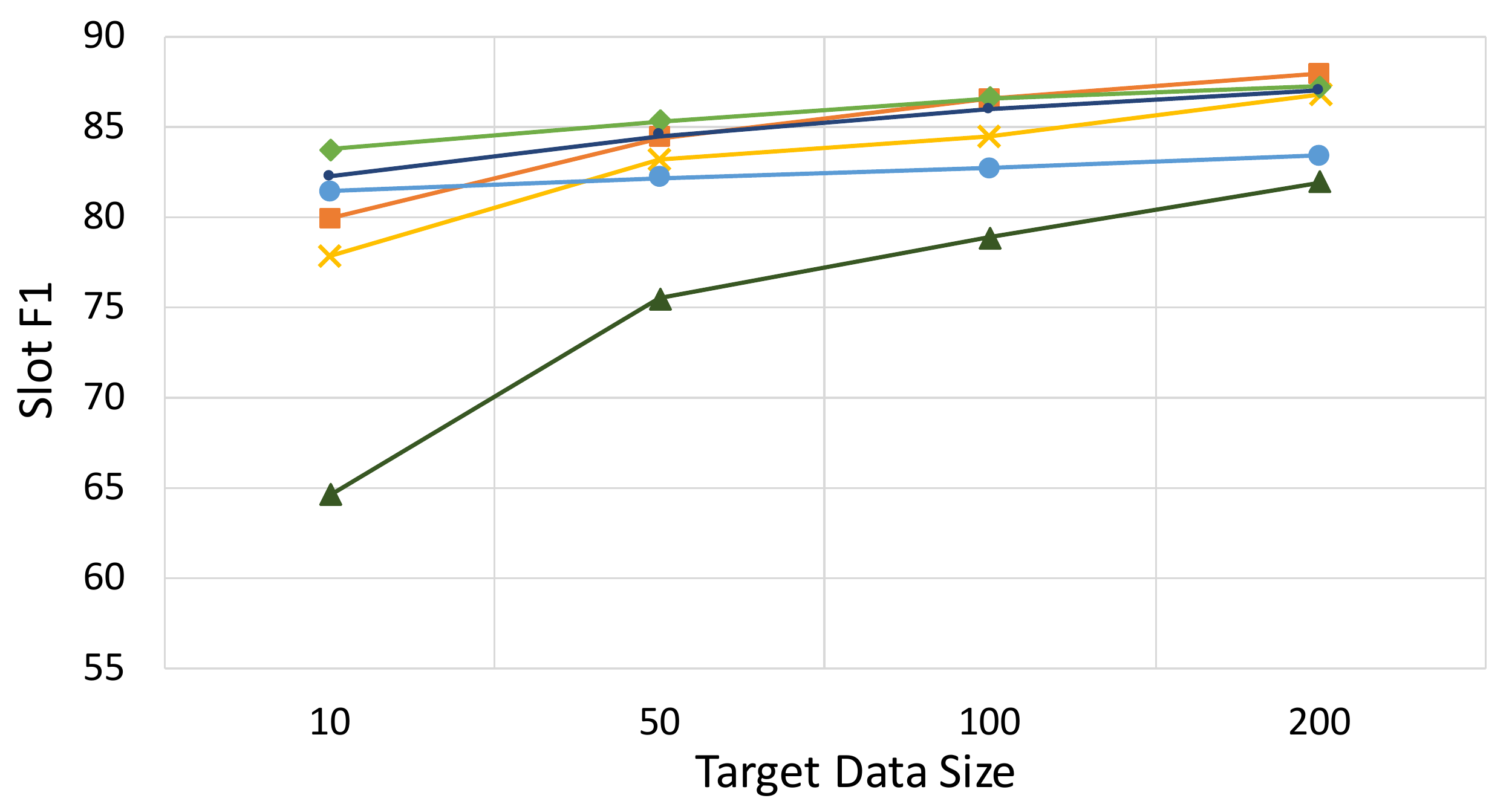} }}\hspace{1em}
    \subfloat[Slot F1 on Chinese]{{\includegraphics[width=0.46\textwidth]{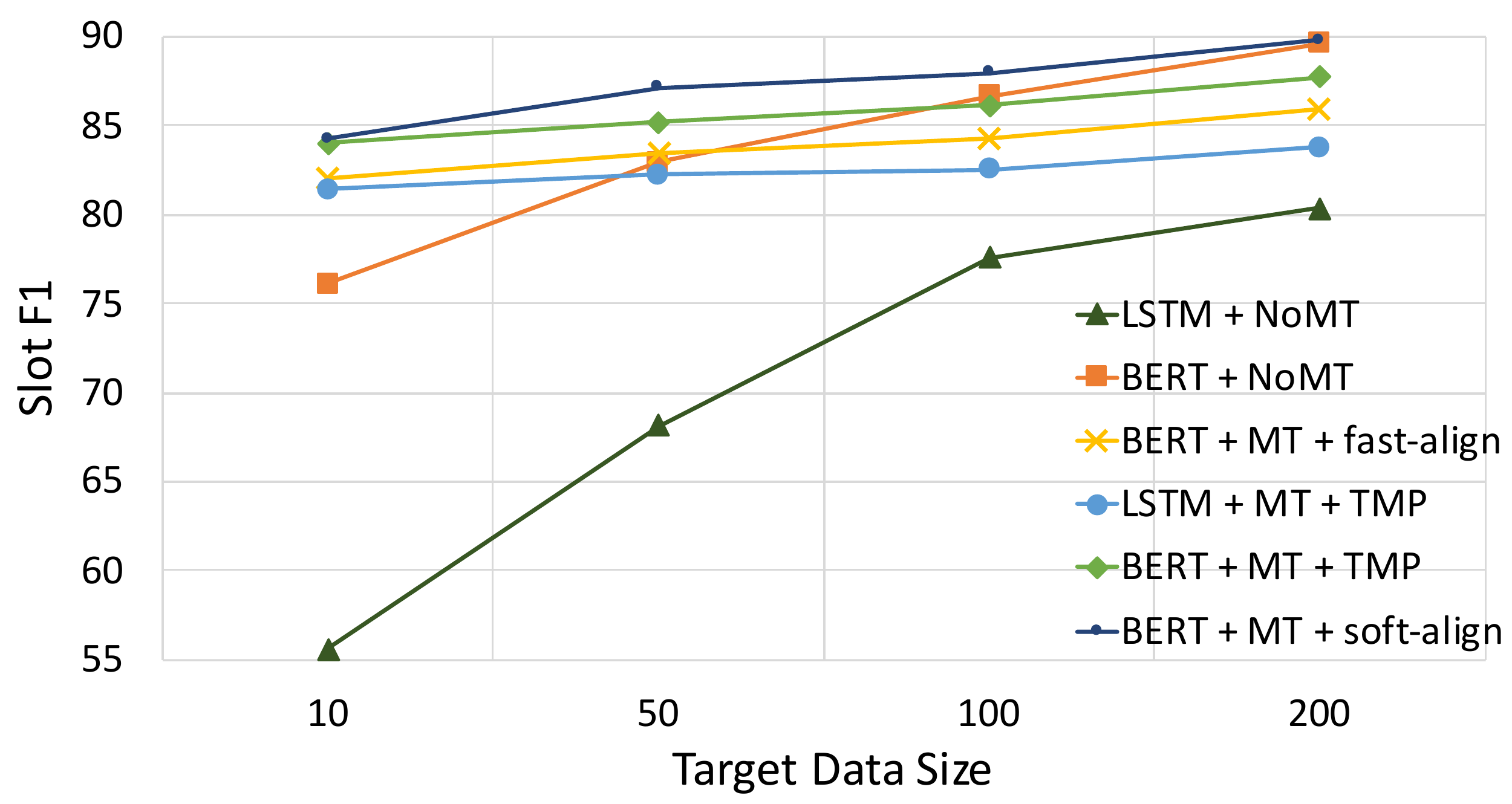} }}
\caption{Results for cross-lingual transfer given various sizes of the labeled data from the target language in addition to the English and machine translated data used in zero-shot experiments. We report scores averaged over 5 runs using target training data selected with different random seeds. Results show the effectiveness of our soft-alignment method as compared to the best projection-based method~(\textit{BERT+MT+TMP}) in few-shot transfer where only a few hundred target examples are available.}
	\label{fig:learning_curve}
\end{figure*}

\looseness=-1
Next, we compare our soft-alignment method with other MT-based approaches. Our method outperforms \textit{MT+fast-align} on five of the eight languages and is more robust across languages \---\ it achieves consistent improvements over BERT fine-tuned without MT on both intent accuracy and slot F1, while \textit{MT+fast-align} leads to a degradation on French and Turkish. 
Furthermore, we compare our method with \textit{MT+TMP}, a strong label projection baseline that combines MT and linguistic features to improve the quality of label projection.
Results show that our method uses only half the training time as \textit{MT+TMP}~(Table~\ref{tab:zeroshot_runtime}), while achieving competitive performance~(Table~\ref{tab:zero_results}): on intent accuracy, our method achieves on par or higher scores on all languages. On slot F1, our method performs on par or better than \textit{MT+TMP} on four of the eight languages and achieves close performance on the remaining languages. 

Finally, these results emphasize the need to evaluate cross-lingual transfer methods on a diverse set of languages to fully illustrate the strengths and weaknesses of each method and to avoid overly strong claims. In our experiments, we find that although \textit{MT+TMP} achieves higher slot F1 than our method on some languages, our method is more robust across languages and outperforms \textit{MT+TMP} by a large margin~(+16.9\%) on Turkish. To further explain the large gap, we measure the slot projection accuracy of \textit{fast-align} and \textit{TMP} compared against the annotated data. We find that \textit{fast-align} obtains extremely low projection accuracy~(20\%) on Turkish, and \textit{TMP} improves it to~39\%, which is still low compared to that on other languages~(above~70\%). The low projection accuracy on Turkish can be attributed to the morphological difference between Turkish and English, which makes it difficult to hard-align each Turkish word to a single English word.

\begin{table}[t]
\centering
\begin{tabular}{lcccccccc}
\toprule
{\bf Intent acc.} & average \\
\midrule
{MT+soft-align} & {94.87} \\
{w/o reconstruction loss} & {94.95} \\
{w/o joint training on source} & {94.64} \\
\bottomrule
{\bf Slot F1} & average \\
\midrule
{MT+soft-align} & {80.00} \\
{w/o reconstruction loss} & {76.42} \\
{w/o joint training on source} & {71.62} \\
\bottomrule
\end{tabular}
\caption{Ablation results for zero-shot transfer learning. Scores are averaged over Spanish, German, Chinese, Japanese, Portuguese, French, Hindi, and Turkish. Ablating the reconstruction loss or joint training on the source language data using supervised objective leads to a major drop on slot F1, while the impact on intent accuracy is small.}
\label{tab:ablation}
\end{table}

\subsection{Learning Curves}
\looseness=-1
Figure~\ref{fig:learning_curve} shows the few-shot transfer results where we add a small amount of labeled data from the target language.\footnote{We apply only the standard supervised objective on the target language data.} We select French as a similar language to English, and Chinese as a dissimilar language. We find that \textit{BERT+NoMT} obtains promising results in the few-shot setting with only several hundred training examples in the target language \---\ it achieves comparable or even higher scores than the best MT-based approach on French, but still lags behind on Chinese by around 5\% on intent accuracy. In addition, results show the effectiveness of our method in few-shot transfer: our method obtains comparable~(on French) or higher~(on Chinese) slot F1 than the best projection-based method~(\textit{BERT+MT+TMP}) given a few hundred target examples, which suggests that, even with a less noisy label projector, the projection errors may still hinder the model from best exploiting the small amount of target language data especially on languages dissimilar to English.

\subsection{Ablation Study}
We evaluate the impact of different components in our soft-alignment model. Table~\ref{tab:ablation} shows that both the reconstruction loss and joint training on the source data using supervised objective are beneficial \---\ slot F1 drops by~3.6\% when ablating the reconstruction loss and by~8.4\% when ablating joint training on the source, while both have little impact on intent accuracy.

\section{Conclusion}
We introduce MultiATIS++, a multilingual NLU corpus that extends the Multilingual ATIS corpus to nine languages across four language families. We use our corpus to evaluate various cross-lingual transfer methods including the use of multilingual BERT encoder, machine translation, and label projection. We further introduce a novel end-to-end model for joint slot label alignment and recognition that requires no external label projection. Experiments show that multilingual BERT brings substantial improvements on multilingual training and cross-lingual transfer tasks.
Furthermore, our model outperforms the simple projection baseline using fast-align on most languages, and achieves competitive performance to the state-of-the-art label projection approach with only half of the training time. 
We release our MultiATIS++ corpus to facilitate future research on cross-lingual NLU to bridge the gap between cross-lingual transfer and supervised methods.

\section*{Acknowledgments}
\looseness=-1
We thank Tamim Swaid and Olga Pospelova at the Amazon AI Data team for collecting the data, the anonymous reviewers, Eleftheria Briakou, Marine Carpuat, Pranav Goel, Pedro Rodriguez, the CLIP lab at UMD, and the Amazon AI team for their valuable feedback.

\bibliography{anthology,emnlp2020}
\bibliographystyle{acl_natbib}

\clearpage
\appendix
\section{Model and Training Details}
\label{appendix:model}
We train all models on 4 NVIDIA V100 Tensor Core GPUs. For both the multilingual NLU and cross-lingual transfer experiments, we train the models using the Adam optimizer~\cite{Kingma2015} for 20 epochs. We set the initial learning rate to~$10^{-3}$ for the LSTM model and~$10^{-5}$ for the BERT model. The LSTM model has embeddings of size 256 and 128 hidden units. We add dropout of~0.1 to the embeddings and encoder hidden states. Both LSTM and BERT models use the WordPiece tokenization model from \cite{Devlin2019}. Table~\ref{tab:num_params} shows the total number of parameters for each model, and Table~\ref{tab:multi_runtime} shows the training time used for each method in the multilingual NLU experiments.
\begin{table}[h]
\centering
\begin{tabular}{lr}
\toprule
& {\#params} \\
\midrule
{LSTM} & $27,539,866$  \\
{BERT} & $166,884,250$ \\
{BERT+soft-align} & $169,944,625$ \\
\bottomrule
\end{tabular}
\caption{Total number of parameters for each model. Our model contains additional parameters for the attention module and linear output layer for the reconstruction loss.}
\label{tab:num_params}
\end{table}

\begin{table}[h]
\centering
\begin{tabular}{llr}
\toprule
\multicolumn{2}{l}{} & {Time (mins)} \\
\midrule
\multirow{2}{*}{Target only} & {LSTM} & 18  \\
{} & {BERT} & 65 \\
\multirow{2}{*}{Multilingual} & {LSTM} & 9 \\
{} & {BERT} & 51  \\
\bottomrule
\end{tabular}
\caption{Total training time for all languages in the multilingual NLU experiments.}
\label{tab:multi_runtime}
\end{table}

\section{Evaluation}
We evaluate all NLU models using intent accuracy and slot F1. Before computing slot F1, we merge all slots that are segmented during preprocessing to match with the original slot segments. We use the script \verb|conlleval.pl|\footnote{\url{http://deeplearning.net/tutorial/code/conlleval.pl}} to compute slot F1.

\section{Validation Performance}
We report the average intent accuracy and slot F1 on the development sets in the multilingual NLU experiments in Table~\ref{tab:valid_results}. For zero-shot experiments, we select the model at the last epoch as we assume no access to the development sets from the target language in this setting.

\begin{table}[h]
\centering
\begin{tabular}{llrrrrrrrrr}
\toprule
\multicolumn{2}{l}{\bf Intent acc.} & {Average} \\
\midrule
\multirow{2}{*}{Target only} & {LSTM} & 94.98 \\
{} & {BERT} & 97.12 \\
\multirow{2}{*}{Multilingual} & {LSTM} & 98.57 \\
{} & {BERT} & 98.37 \\
\bottomrule
\multicolumn{2}{l}{\bf Slot F1} & {Average} \\
\midrule
\multirow{2}{*}{Target only} & {LSTM} & 89.41 \\
{} & {BERT} & 91.91 \\
\multirow{2}{*}{Multilingual} & {LSTM} & 97.64 \\
{} & {BERT} & 98.82 \\
\bottomrule
\end{tabular}
\caption{Average intent accuracy and slot F1 on the development sets in the multilingual NLU experiments. For \textit{targe only} models, we average the scores over all nine languages. For \textit{multilingual} models, we only validate on the English development set.}
\label{tab:valid_results}
\end{table}
\end{document}